\documentclass[letterpaper, 10 pt, conference]{ieeeconf}  

\IEEEoverridecommandlockouts                              

\usepackage{amsmath}
\usepackage{amsfonts}
\usepackage{bm}
\usepackage{graphicx}
\usepackage{siunitx}
\usepackage{hyperref}
\usepackage{acro}
\usepackage{xspace}
\usepackage{balance}

\usepackage{cite}

\DeclareAcronym{jit}{short = JIT, long = just-in-time}
\DeclareAcronym{bem}{short = BEM, long = Blade Element Momentum}
\DeclareAcronym{rpy}{short = rpy, long = roll-pitch-yaw}
\DeclareAcronym{mpc}{short = MPC, long = model predictive control}
\DeclareAcronym{rl}{short = RL, long = reinforcement learning}
\DeclareAcronym{shac}{short = SHAC, long = short-horizon actor-critic}
\DeclareAcronym{bptt}{short = BPTT, long = backpropagation through time}

\title{\LARGE \bf
UWB Meets \texttt{Crazyflow}: Simulating Degraded Feedback at Scale 
\\for Aerial Robotics
}

\author{Martin Schuck$^{*}$, Marcel P. Rath$^{*}$, Yufei Hua, Abhishek Goudar, SiQi Zhou, Angela P. Schoellig
\thanks{*Equal contribution. The authors, Martin Schuck, Marcel Rath, Yufei Hua, and Angela P. Schoellig, are with the \href{https://www.learnsyslab.org/}{Learning Systems and Robotics Lab}, Technical University of Munich, Germany. Abhishek Goudar is with the University of Toronto, Canada. SiQi Zhou is with Simon Fraser University, Canada.}
\thanks{This work was supported by the \href{https://www.robotics-institute-germany.de/}{Robotics Institute Germany} under BMBF grant 16ME0997K and by the Humboldt Professorship for Robotics and Artificial Intelligence.}
}

\begin{document}

\maketitle
\thispagestyle{empty}
\pagestyle{empty}

\begin{abstract}
In this work, we introduce \texttt{Crazyflow}, an accurate, differentiable simulator built on JAX. By leveraging \ac{jit} compilation via XLA, \texttt{Crazyflow} unifies physics and control into a single differentiable computation graph, enabling massive parallelization on accelerated hardware without sacrificing modeling accuracy. This architecture achieves order-of-magnitude speedups over existing baselines, capable of training deployable reinforcement learning agents in seconds. To highlight its highly modular design, we demonstrate how easily \texttt{Crazyflow} can be extended by integrating a complete, high-fidelity Ultra-Wideband (UWB) and Inertial Measurement Unit (IMU) simulation pipeline coupled with a full-state Extended Kalman Filter (EKF). This capability allows for massive parallel controller evaluation under realistic, degraded state feedback with minimal impact on GPU throughput. By combining speed, accuracy, and extensibility, \texttt{Crazyflow} serves as a foundational tool for the next generation of aerial robotics research.
\end{abstract}

\section{Introduction}
Owing to their high agility, mechanical simplicity, and cost-effective design, quadrotors have emerged as a canonical platform across a broad spectrum of robotics applications, from infrastructure inspection and safety-critical tasks~\cite{lyu2023unmanned,guan2022review} to recreational domains such as racing and shows~\cite{hanover2024autonomous,schoellig2014so}. Advancing capabilities in either pushing individual robots toward their physical limits or scaling up multi-agent systems necessitates high-performance, efficient simulation tools that enable rapid algorithmic iteration and task optimization.

Widely used simulators such as \texttt{RotorS}~\cite{furrer2016rotors}, \texttt{CrazyS}~\cite{silano2018crazys}, and swarm-oriented frameworks like \texttt{Crazyswarm}~\cite{preiss2017crazyswarm} and \texttt{CrazySim}~\cite{llanes2024crazysim} are built on top of modular frameworks and provide accurate models suitable for controller development. However, evaluating controllers under imperfect, noisy state feedback—such as realistic UWB or vision pipelines—across thousands of parallel rollouts remains a significant scaling bottleneck, as these tools are generally not designed for massively parallel GPU execution.

To address the computational demands of large-scale learning, frameworks like \texttt{PRL4AirSim}~\cite{saunders2023parallel}, \texttt{gym-pybullet-drones}~\cite{panerati2021learning}, and \texttt{Aerial Gym}~\cite{kulkarni2025aerialgym} leverage distributed or tensorized physics to achieve high throughput. While highly scalable, they typically do not expose analytical gradients of the underlying dynamics, which are required by a range of advanced learning- and optimization-oriented methods. Emerging differentiable frameworks attempt to bridge this gap~\cite{zhang2025visionbased,zhang2025diffaero}, but often compromise on model fidelity or lack demonstrated sim-to-real capabilities outside of simplified settings.

\begin{figure}
    \centering
    \includegraphics[width=\columnwidth]{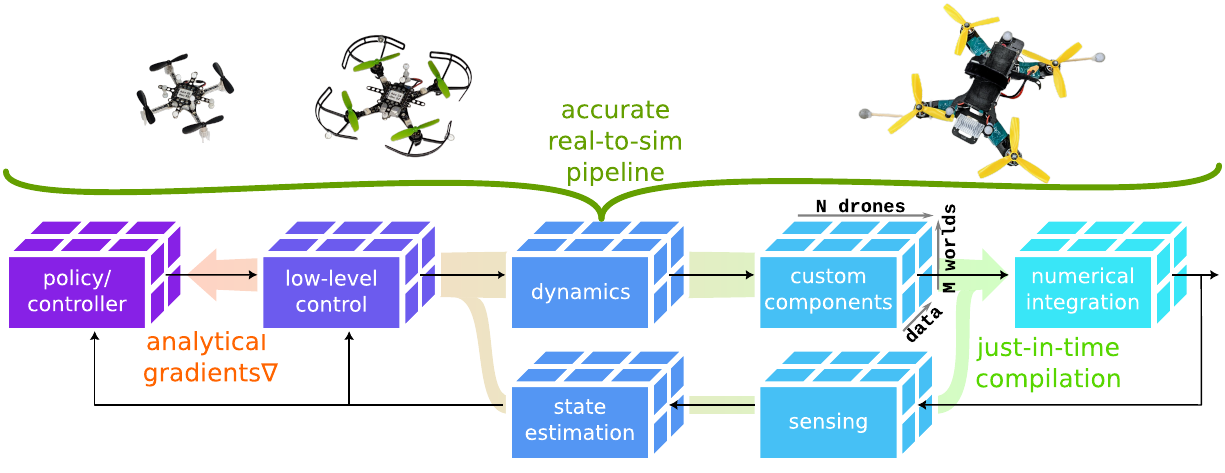}
    \caption{An overview of the \texttt{Crazyflow} simulator. Built on JAX, \texttt{Crazyflow} leverages just-in-time compilation via XLA to unify physics, sensing, estimation, and controllers into a single differentiable computation graph. Website: \href{https://learnsyslab.github.io/crazyflow}{\url{https://learnsyslab.github.io/crazyflow/}}.}
    \label{fig:overview}
\end{figure}

In this work, we introduce \texttt{Crazyflow} (Fig.~\ref{fig:overview}), a high-fidelity, highly extensible, differentiable simulator built on JAX. \texttt{Crazyflow} unifies physics, modular sensor pipelines, and controllers into a single computation graph. To demonstrate this robust capability, we seamlessly integrate a UWB and IMU sensing pipeline with an Extended Kalman Filter (EKF). This allows for realistic controller evaluation under imperfect state estimation at massive scale without sacrificing the core tenets of the simulator: high-speed, direct sim-to-real transfer, and full differentiability.

\section{The Crazyflow Design} 
\label{sec:crayzflow}

\texttt{Crazyflow} leverages JAX to deliver \ac{jit} compilation, massively parallel hardware acceleration, and full differentiability. We dynamically compose the simulation pipeline using only the functions required for the current configuration. By focusing on unconstrained free-flight, we omit the computationally expensive solver step found in general-purpose simulators, simplifying gradient tracing and significantly enhancing performance. 

\subsection{Core Dynamics and Sim-to-Real}
Modeled natively after the Crazyflie quadrotor family, \texttt{Crazyflow} provides two highly accurate dynamics models: a detailed first-principles model~\cite{forster2015system, luis2016design} capturing non-linear motor dynamics, and an abstracted attitude dynamics model designed for standard mid-level control interfaces. Beyond direct compatibility with Crazyflies, \texttt{Crazyflow} incorporates a streamlined system identification pipeline. This allows users to effortlessly adapt the simulation framework to extract and match the physical parameters of entirely new drone platforms, guaranteeing reliable sim-to-real transfer regardless of the hardware.

\subsection{Extensibility: Sensing and Estimation}
Because \texttt{Crazyflow} is built on a foundation of stateless, pure functions, the architecture permits extensive customization. To highlight this modularity, we integrated a complete sensing and state-estimation pipeline directly into the simulator step execution ahead of the controller. 

The sensor models feature an IMU and a UWB network. The IMU simulates body-frame specific force and angular velocity, corrupted by additive white noise and slowly varying bias random walks~\cite{trawny2005indirect}. 
The UWB model simulates one tag on the drone interacting with eight fixed anchors located at $x, y \in \{-2.5, 2.5\}\,\si{\meter}$ and $z \in \{0, 3\}\,\si{\meter}$. 
It outputs the true Euclidean distance corrupted by Gaussian noise with a fixed standard deviation $\sigma$ and a mean $\mu_i$. Assuming line-of-sight (LoS) conditions throughout the flight, this mean is uniformly sampled at initialization per anchor $i$ and held constant per episode to reflect realistic hardware offsets~\cite{ledergerber2015robot}.

These measurements are fused via a full-state EKF that tracks 19 states: position, quaternion attitude, linear velocity, angular velocity, and the respective accelerometer and gyroscope biases. The prediction step propagates the mean through a nonlinear IMU-driven dynamics model and propagates covariance using first-order linearization via Jacobians. The correction step uses the UWB range measurement model and its linearization with respect to position. This allows users to seamlessly switch the controller input source between perfect ground-truth state and realistic estimated feedback.

\section{Evaluation Results} \label{sec:results}

\begin{figure}
    \centering
    \includegraphics[width=0.49\textwidth]{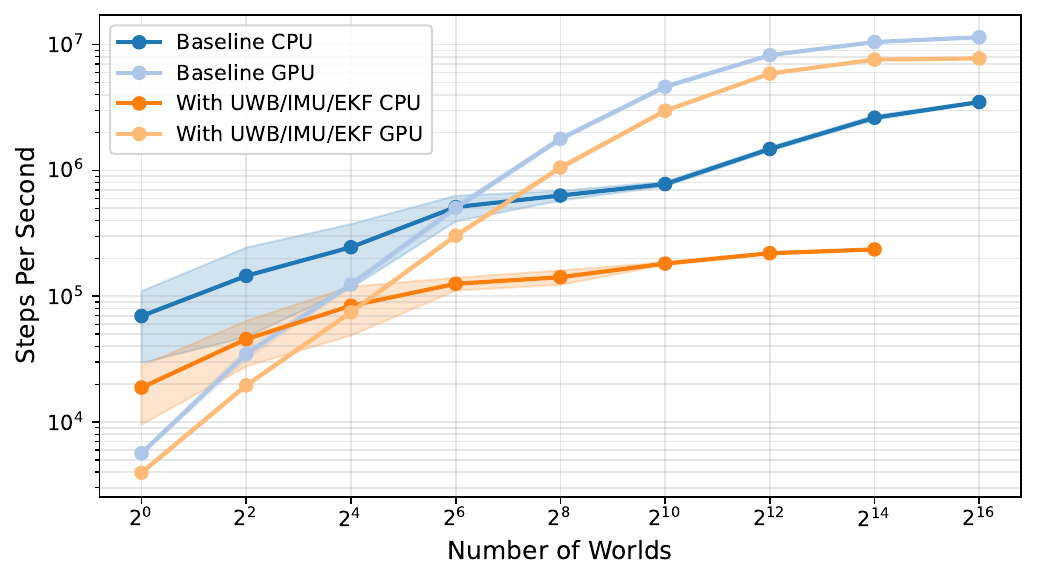}
    \vspace{-2em}
    \caption{Steps per second for a single-drone simulation in \texttt{Crazyflow} with and without a full state estimation pipeline integrated. GPU accelerated results are depicted in lighter shades.}
    \label{fig:performance}
    \vspace{-0.5em}
\end{figure}

First, we evaluate the computational throughput of \texttt{Crazyflow} with and without the integrated state estimation pipeline. Benchmarks were conducted on a workstation equipped with an NVIDIA RTX 5090 GPU and an Intel Core Ultra 9 285K CPU. As shown in Fig.~\ref{fig:performance}, integrating the complex matrix inversions and linearizations of the 19-state EKF introduces some computational overhead. While this impact is quite noticeable on the CPU backend, the GPU simulation handles the load exceptionally well. The GPU experiences only a roughly $20\%$ throughput reduction across all parallel world counts, maintaining a peak performance of ${\sim}8 \times 10^6$ steps per second, compared to slightly over $10^7$ for the baseline. This demonstrates that degraded feedback simulation can be run without breaking scalability.

\begin{figure}
    \centering
    \includegraphics[width=0.49\textwidth]{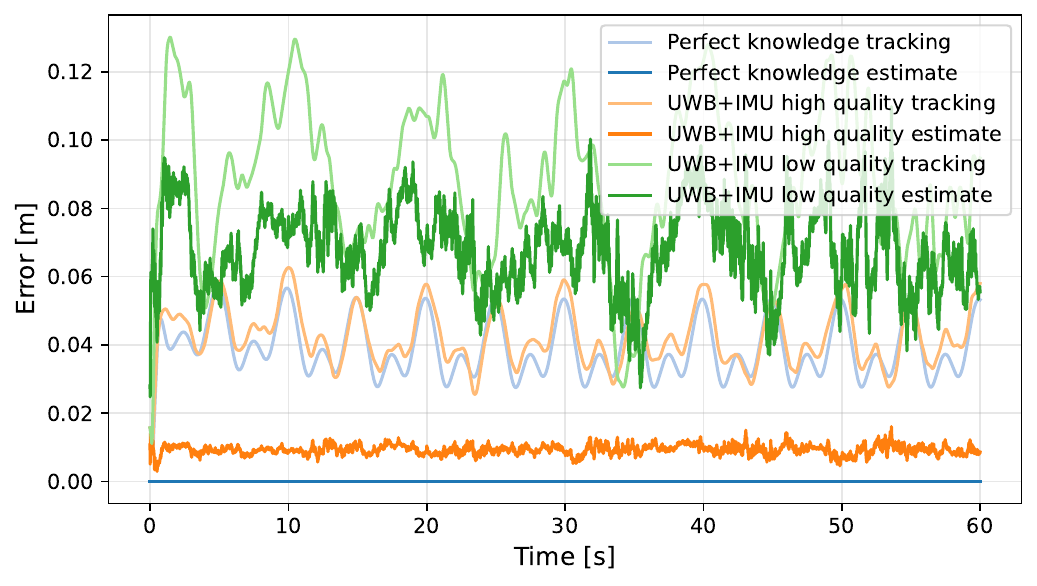}
    \vspace{-2em}
    \caption{Simulated tracking of a figure-8 reference with a \SI{10}{\second} cycle time using a geometric controller \cite{mellinger2011controller}. With lower sensor quality, the position estimate and the tracking quality degrade.}
    \label{fig:policy}
    \vspace{-1.0em}
\end{figure}

Second, we evaluate closed-loop trajectory tracking under varying sensor configurations. A geometric controller~\cite{mellinger2011controller} tracks a figure-8 reference trajectory with a \SI{10}{\second} cycle time. 
We compare three distinct conditions: perfect ground-truth state feedback, a high-quality UWB+IMU configuration ($\mu_i\sim\mathcal{U}([0,\SI{0.02}{\meter}])$, $\sigma = \SI{0.01}{\meter}$), and a low-quality configuration ($\mu_i\sim\mathcal{U}([0,\SI{0.15}{\meter}])$, $\sigma = \SI{0.10}{\meter}$).
As depicted in Fig.~\ref{fig:policy}, degrading the sensor quality directly degrades the EKF's position estimate, which correspondingly deteriorates the overall closed-loop tracking performance.

Crucially, despite the addition of realistic sensor noise, bias sampling, and EKF linearizations, the entire pipeline remains completely differentiable. This preserves \texttt{Crazyflow}'s most powerful feature: using analytical gradients via \ac{bptt}, \ac{rl} agents can still be trained in seconds and successfully deployed directly on physical hardware without relying on complex domain randomization. Users can now design, train, and robustify advanced policies against imperfect state feedback directly within the learning loop.

\section{Conclusion}
We introduced \texttt{Crazyflow}, an accurate, JAX-based simulation environment designed to bridge the gap between execution speed, full differentiability, and physical realism. \texttt{Crazyflow} allows researchers to adapt any drone platform via an integrated system identification pipeline and enables the training of deployable \ac{rl} agents in seconds, supporting reliable, direct sim-to-real transfer. By effortlessly integrating a full UWB/IMU and EKF pipeline into the step execution with minimal GPU overhead, we demonstrated the simulator's capability to evaluate robust control under realistic, imperfect state feedback at scale. Combining fast execution with high-fidelity sensory realism, \texttt{Crazyflow} establishes itself as a foundational tool to push the boundaries of next-generation aerial robotics research.

\clearpage
\balance
\bibliographystyle{IEEEtran}
\bibliography{references}

\end{document}